\documentclass[11pt]{article}
\usepackage[margin=1in]{geometry}
\usepackage{graphicx}
\usepackage[lofdepth,lotdepth]{subfig}
\usepackage{placeins}

\usepackage[normalem]{ulem}
\usepackage{tikz-cd}

\usepackage{adjustbox}

\usepackage{amsmath,amssymb}
\usepackage{hyperref}
\hypersetup{
colorlinks = true,
linkcolor = blue,
citecolor = blue,
urlcolor = blue
}

\title{A Note on Argumentative Topology: Circularity and Syllogisms as Unsolved Problems}

\author{Wlodek W. \ Zadrozny$^{1,2,}$ \\
wzadrozn@uncc.edu\\
$^1$ College of Computing, University of North Carolina at Charlotte\\
$^2$ School of Data Science, University of North Carolina at Charlotte}
\date{January 05-25, 2021} 

\begin{document}

\maketitle

\begin{abstract}
In the last couple of years there were a few attempts to apply topological data analysis to text, and in particular to natural language inference. A recent  work by Tymochko et al. suggests the possibility of capturing `the notion of logical shape in text,'  using `topological delay embeddings,' a technique derived from dynamical systems, applied to word embeddings. 

In this note we reconstruct their argument and show, using several old and new examples, that the problem of connecting logic, topology and text is still very much unsolved.  We conclude that there is no clear answer to the question:  ``Can we find a circle in a circular argument?'' We point out some possible avenues of exploration. The code used in our experiment is also shown.

\end{abstract}

\section{Introduction}\label{sec:intro}

This note describes our attempt to reconstruct the recent work by Tymochko et al.  \cite{tymochko2020argumentative}, suggesting the possibility of capturing `the notion of logical shape in text,'  using `topological delay embeddings,' a technique derived from dynamical systems, and applied to word embeddings. The authors argue that using topological techniques it might be possible to ``find a circle in a circular argument?''

The authors say \cite{tymochko2020argumentative}:
\begin{quote}
We were originally motivated by the question ``Why do we call a circular argument `circular'?'' A circular argument is one that logically loops back on itself. This intuitive definition of why the argument is circular actually has an analogous and mathematically precise definition from a topological perspective. Topology is the mathematical notion of shape and a topological circle can be abstractly defined as any shape which starts and loops back on itself (i.e. a circle, a square, and a triangle are all topologically a circle). 
\end{quote}

This an interesting problem, and to address it, the cited work uses a new technique of \textit{Topological Word Embeddings} (TWE). The method combines topological data analysis (TDA) with mathematical techniques in dynamical systems, namely time-delayed embeddings. The cited article suggests a positive answer to the motivating question. 

However, in this note, we argue, using similar examples, that the process of finding circularity using TDA might be more complicated. We describe our attempt, after reconstructing the TWE method, in  applying it to circular and non-circular examples. We observe that the method often fails to distinguish the two cases. Thus, while in the cited work \cite{tymochko2020argumentative} and {elsewhere \cite{savle2019topological} }, we have some interesting examples connecting topological data analysis and inference, the problem of connecting reasoning with topological data analysis is, in our opinion, still open. 

This note is organized as follows. We dispense with the preliminaries, and instead we refer the reader who is not familiar with topological data analysis to the original article \cite{tymochko2020argumentative} for the TDA background and the description of the experiment which motivated this reconstruction and extension. Similarly, since word embeddings are a standard natural language representation technique, we only need to mention that it consists in replacing words (or terms) by 50 to 1000 dimensional vectors, which have been proven to capture e.g. words similarities. Again we refer the reader to the standard sources e.g. \cite{mikolov2013distributed,pennington2014glove}. Finally, we make no attempt to put this note into a broader context of TDA-based research, and we provide only a bare minimum of references. \\

In Section \ref{sec:methodAndSettings} we briefly describe the method, as well as the software and settings used in our experiment. Section \ref{sec:examples} is all about examples showing the \textit{instability} of the TWE method; that is a slight change in parameters results in markedly different persistence diagram. We finish, in Section~\ref{sec:discussion},
with a discussion and propose some avenues of research on connecting TDA and inference. The Appendix has the code we used for these experiments. 

\section{Method and software used in the experiment}\label{sec:methodAndSettings}

The purpose of out experiment was to see whether circularity of arguments corresponds to the presence of loops in a representation of the sentences in the arguments as reported in \cite{tymochko2020argumentative}.
Briefly, the representation to which TDA is applied is created by the following steps:

\begin{quote}
\begin{enumerate}
\item The words in the sentences are replaced by their embedding vectors 

\item Since the words occur in a sequence, the embedding vectors can be viewed as a series of vectors. 

\item To get a one-dimensional time series we compute the dot product of each vector with a random vector of the same dimension. 

\item This time series is then processed using a time delay embedding method

\item Finally, the TDA is applied and the persistence computed and displayed.

\end{enumerate}
\end{quote}

These are exactly the steps in Fig. 1 in \cite{tymochko2020argumentative}. However, we need to clarify some aspects of our experiment, since the original paper does not mention certain details. The following list refers to the steps above: 

\begin{quote}
\begin{enumerate}
\item We used 50, 100, 200 and 300 dimensional Glove embedding vectors \cite{pennington2014glove}. Glove vectors were also used in \cite{tymochko2020argumentative}.

\item (nothing to add here)  

\item We used the \texttt{numpy} dot product, and random seed of 42 in most experiments, but we also used the seeds of 1, 2 and 3 (to show that changing the seed changes the persistence diagrams.

\item We used the function \texttt{takensEmbedding } from the public repository  available on Kaggle (\small{\url{https://www.kaggle.com/tigurius/introduction-to-taken-s-embedding}}) \
\normalfont
It implements the Takens  time delay embedding method \cite{Takens:81}. However, we did not investigate the optimal parameters, and instead we used the time delay of 2, and dimension 2, as in the original paper. The cited Kaggle code allows us to search for optimal parameters, and we intend to use it in further experiments. In this note, we also looked into into (2,3) and (3,2) as time delay and dimension of Takens embeddings, and observed such changes can significantly impact the patterns shown in persistence diagrams. 

\item For TDA and persistence we used the Python version (0.6) 
\small{\url{https://ripser.scikit-tda.org/en/latest/}}
\normalfont
 of Ripser \cite{ripser}.

\end{enumerate}
\end{quote}

\section{Examples of circular and not circular reasoning}\label{sec:examples}

In this section we first look at persistence diagram for circular reasoning. We compare them with examples with very similar words and structures which are not circular. Then we perform the same exercise for syllogisms. 
We use two examples are from the cited paper. The other ones are ours, and are intended to test the influence of changes in vocabulary, paraphrase, and patterns of reasoning. 

\begin{itemize}

\item Circular (\cite{tymochko2020argumentative}): \textit{ “There is no way they can win because they do
not have enough support. They do not have enough support because there is no way they
can win.”}

\item Circular (ours, word substitution to the previous one):  \textit{``There is no way for the crew to win because the crew does not have good rowers. The crew does not have good rowers because there is no way for the crew to win."}

\item Circular (ours, modified paraphrase of the above ): \textit{``The Russian crew must lose  because the coach did not hire Siberian rowers. The team did not enlist the good Siberian rowers because there is no way for the Russian crew to win." }

\item  Non-circular (\cite{tymochko2020argumentative}): \textit{“There is no way they can win if they do not have enough support. They do not have enough support, so there is no way they can win.”}

\item Non-circular (ours): \textit{ "No way the anarchists can lose the primary elections if they have enough support. The anarchists have strong support, so there is no way they can lose the primaries."}

\item Inductive reasoning (ours): \textit{``Gold is going up. Platinum is also going up. We should buy all metals, including copper and silver."}

\item Syllogism (ours): \textit{``Every animal is created by evolution. The lion is an animal. The lion is created by evolution."}

\item Absurd (random text, \cite{tymochko2020argumentative}): \textit{``The body may perhaps compensates for the loss of a true metaphysics. Yeah, I think it's a good environment for learning English. Wednesday is hump day, but has anyone asked the camel..."}

\end{itemize}

\bigskip
Not only do we fail to detect any particular correlation between circularity and persistence, but we also see that persistence, for the same sentences, looks very differently for 50, 100, 200 and 300--dimensional Glove vectors.  In addition, we see the same the patterns for valid reasoning (e.g. syllogism) and for inductive reasoning. Inductive reasoning usually is not logically valid, although depending on the context might be plausible. We might have seen some influence of less frequent words on the persistence diagrams, but we are not sure. 

And contrary to \cite{tymochko2020argumentative}, we see that the absurd, random text cited above can exhibit pretty much any pattern of persistence, depending on the parameters used to find the persistent homology.

The conclusion we will draw from this exercise is that circular, non-circular and absurd modes of argument might exhibit the same visual properties derived from persistence homology, and therefore cannot be distinguished by looking for ``a circle in a circular argument."

\subsection{Circular reasoning}

\vspace{-2ex}

\begin{figure}[ht]
\centering
\hspace*{-3.9em}
\subfloat[Subfigure 1 list of figures text][(g:50; seed:1)]{
\includegraphics[width=0.4\textwidth]{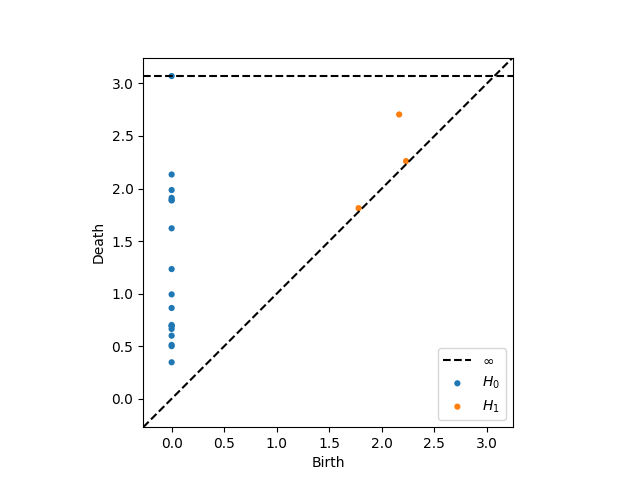}
\label{fig:subfig1cs} \hspace*{-1.9em}}
\subfloat[Subfigure 2 list of figures text][(g:50, seed:2)]{
\includegraphics[width=0.4\textwidth]{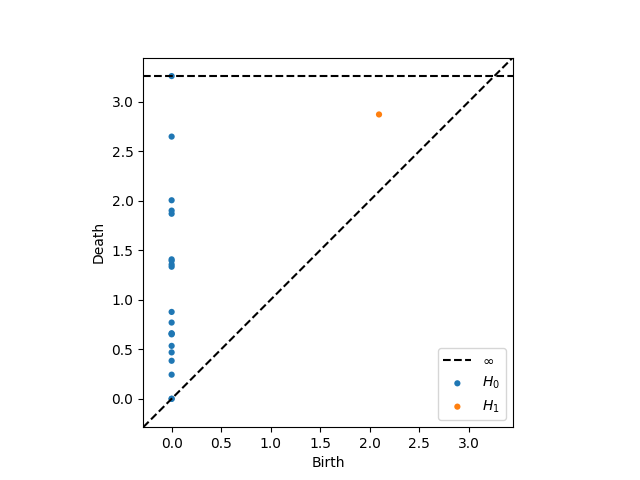}
\label{fig:subfig2cs} \hspace*{-1.9em}}  
\subfloat[Subfigure 3 list of figures text][(g:50; seed:3)]{
\includegraphics[width=0.4\textwidth]{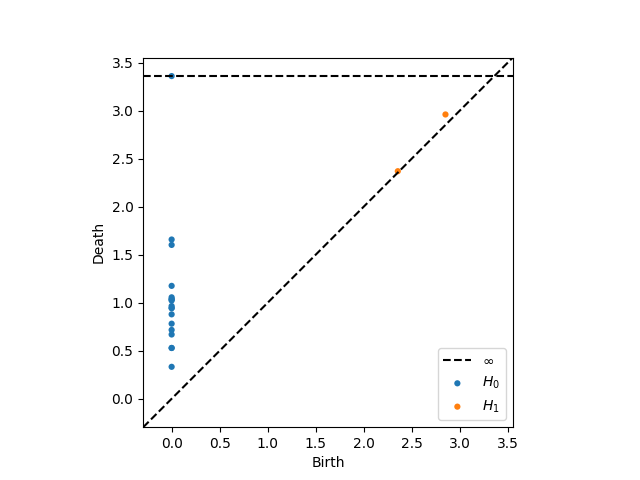}
\label{fig:subfig3cs}}
\caption{\small{Persistence diagrams for the circular text from \cite{tymochko2020argumentative}: \textit{``There is no way they can win because they do
not have enough support. They do not have enough support because there is no way they
can win."}. With different seeds (1,2,3) we get different patterns, even for the same embedding dimension. Here, the dimension of embeddings is 50 (g:50), but the same phenomenon occurs for other dimensions. }}
\label{fig:circSent}
\end{figure}

\vspace{-2ex}

\begin{figure}[hb]
\centering
\subfloat[Subfigure 1 list of figures text][(g:200; seed:42)]{
\hspace*{-2.9em}
\includegraphics[width=0.4\textwidth]{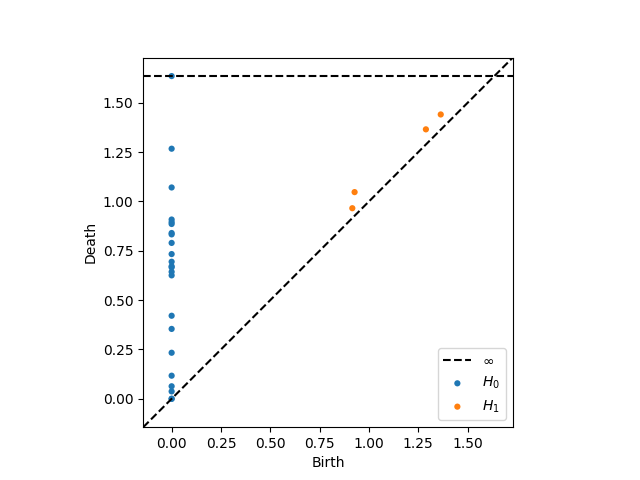}
\label{fig:subfig1csp} \hspace*{-1.9em}}
\subfloat[Subfigure 2 list of figures text][(g:100, seed:42)]{
\includegraphics[width=0.4\textwidth]{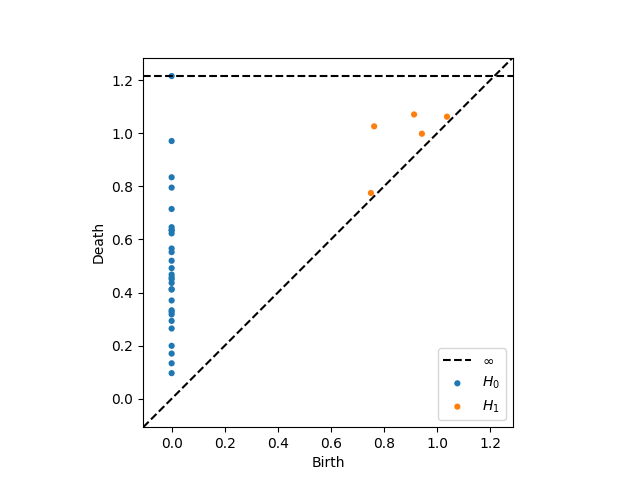}
\label{fig:subfig2csp} \hspace*{-1.9em}}  
\subfloat[Subfigure 3 list of figures text][(g:300; seed:42)]{
\includegraphics[width=0.4\textwidth]{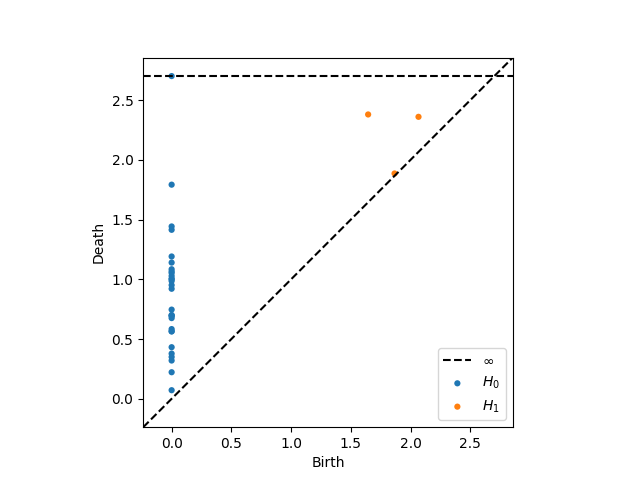}
\label{fig:subfig3csp}}
\caption{\small{Persistence diagrams for two circular text examples. Panel (a) 
\textit{``There is no way for the crew to win because the crew does not have good rowers. The crew does not have good rowers because there is no way for the crew to win."} makes simple vocabulary substitutions to the circular sentences  from \cite{tymochko2020argumentative}. 
Panels (b) and (c) show the patterns obtained with additional changes: \textit{``The Russian crew must lose  because the coach did not hire Siberian rowers. The team did not enlist the good Siberian rowers because there is no way for the Russian crew to win."}}}
\label{fig:circSentp}
\end{figure}

\FloatBarrier

Figure~\ref{fig:circSent} represents persistence diagrams for the circular text example from \cite{tymochko2020argumentative}: \textit{``There is no way they can win because they do
not have enough support. They do not have enough support because there is no way they
can win."}. It shows that with different seeds (1,2,3) we get different patterns, even for the same embedding dimension. In the pictured examples the dimension of embeddings is 50 (g:50), but the same phenomenon occurs for other dimensions. As in \cite{tymochko2020argumentative}, we use the Takens embedding dimension = 2,  and the Takens embedding delay = 2.

Figure~\ref{fig:circSentp} represents persistence diagrams for two circular text examples. Panel (a) makes simple vocabulary substitutions to the circular sentences  from \cite{tymochko2020argumentative}: \textit{``There is no way for the crew to win because the crew does not have good rowers. The crew does not have good rowers because there is no way for the crew to win."}. 
Panels (b) and (c) show the patterns obtained with a few more changes: \textit{``The Russian crew must lose  because the coach did not hire Siberian rowers. The team did not enlist the good Siberian rowers because there is no way for the Russian crew to win." }
These examples show that it is possible to get a `random' pattern by changing vocabulary of the sentences, and such patterns can be `random' or not, depending on the dimension of embeddings.  
As before, we use the Takens embedding dimension = 2,  and the Takens embedding delay = 2.

Based on the examples shown in Figures~\ref{fig:circSent} and \ref{fig:circSentp}, we observe that contrary to the hypothesis proposed in 
 \cite{tymochko2020argumentative}, circular reasoning pattern can produce persistent homology patterns associated with random text. In Section~\ref{sec:randomText} we will see that the opposite is true as well. 

\FloatBarrier

\subsection{Non-circular and syllogistic reasoning}
\vspace{-3ex}
\begin{figure}[h]
\centering
\subfloat[Subfigure 1 list of figures text][Non-circular: (g:200; seed:42)]{
\hspace*{-2.9em}
\includegraphics[width=0.4\textwidth]{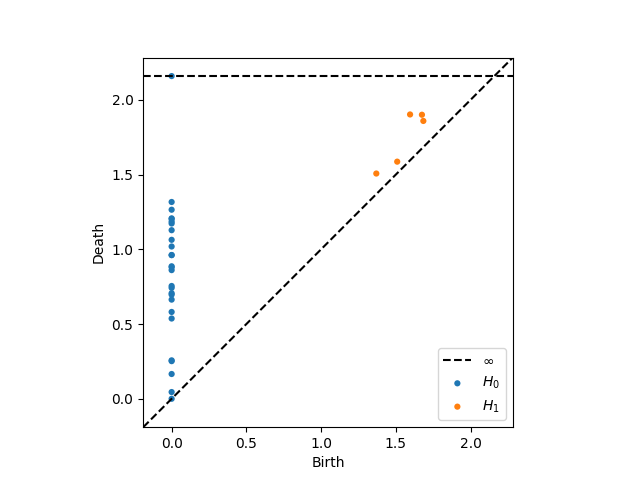}
\label{fig:subfig1nc} \hspace*{-1.9em}}
\subfloat[Subfigure 2 list of figures text][Induction:  (g:50, seed:42)]{
\includegraphics[width=0.4\textwidth]{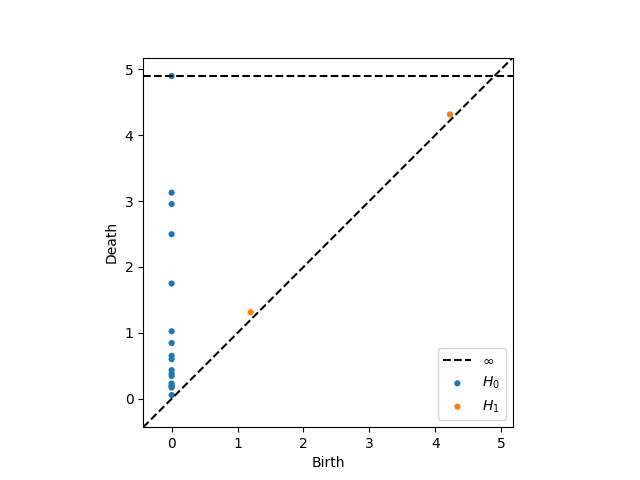}
\label{fig:subfig2nc} \hspace*{-1.9em}}  
\subfloat[Subfigure 3 list of figures text][Syllogism: (g:50; seed:42)]{
\includegraphics[width=0.4\textwidth]{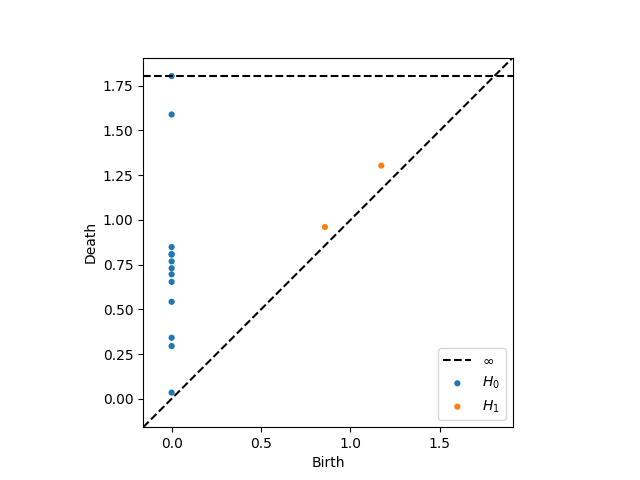}
\label{fig:subfig3nc}}
\caption{\small{Persistence diagrams for three non-circular text examples. Panel (a) represents \textit{ "No way the anarchists can lose the primary elections if they have enough support. The anarchists have strong support, so there is no way they can lose the primaries."} modeled after  \cite{tymochko2020argumentative} non-circular example, but with changed vocabulary and less repetitive pattern. It shows that a valid argument can result in a `random pattern.' 
Panels (b) shows an example of inductive reasoning. 
\textit{'Gold is going up. Platinum is also going up. We should buy all metals, including copper and silver.'}
Panel (c) show the patterns obtained from a syllogism.
\textit{'Every animal is created by evolution. The lion is an animal. The lion is created by evolution.'} 
Even though we see some similarities between (b) and (c), in other dimensions such similarities might not appear. 
As before, we use the Takens embedding dimension = 2,  and the Takens embedding delay = 2.}}
\label{fig:circSentpp}
\end{figure}

Again, despite repeated experiments with different parameters we cannot see any persistent patterns of homological persistence (pun intended). 
The examples show that it is possible to get a `random' pattern by changing vocabulary of non-circular sentences, and such patterns can look `random' or not, depending on the dimension of embeddings.

\noindent
\begin{figure}[]
\centering
\subfloat[Subfigure 1 list of figures text][(g:100, tdim:2, tdel:2)]{
\hspace*{-1.9em}
\includegraphics[width=0.34\textwidth]{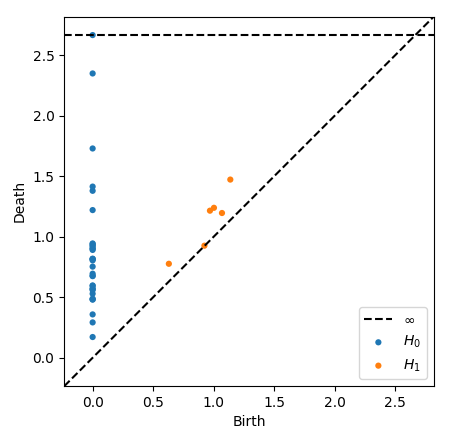}
\label{fig:subfig1r} }
\subfloat[Subfigure 2 list of figures text][(g:200, tdim:2, tdel:2)]{
\includegraphics[width=0.34\textwidth]{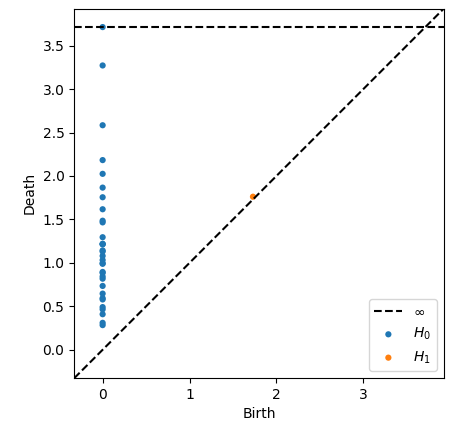}
\label{fig:subfig2r} }
\subfloat[Subfigure 3 list of figures text][(g:300, tdim:2, tdel:2)]{
\includegraphics[width=0.34\textwidth]{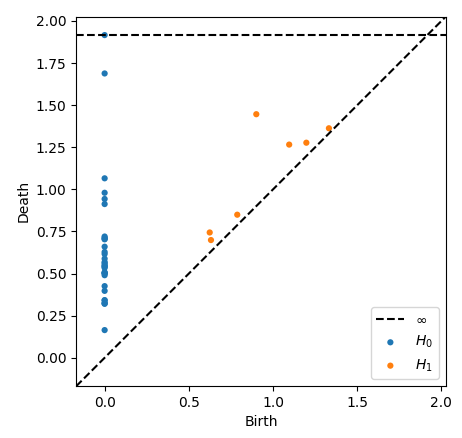}
\label{fig:subfig3r}}\\[-2ex]
\subfloat[Subfigure 4 list of figures text][(g:100, tdim:3, tdel:2)]{
\hspace*{-1.9em}
\includegraphics[width=0.34\textwidth]{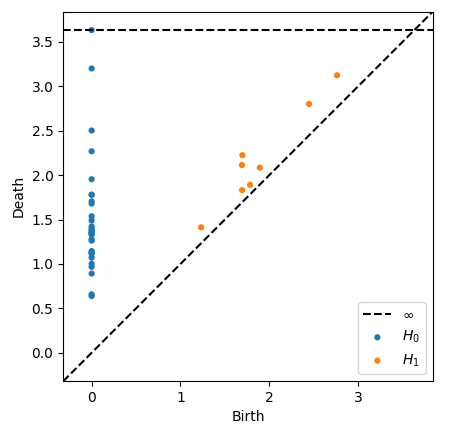}
\label{fig:subfig4r}}
\subfloat[Subfigure 4 list of figures text][(g:200, tdim:3, tdel:2)]{
\includegraphics[width=0.34\textwidth]{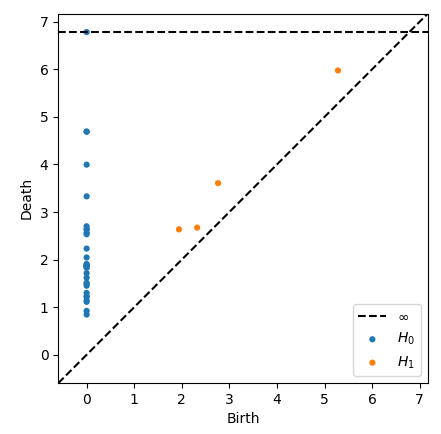}
\label{fig:subfig5r}}
\subfloat[Subfigure 4 list of figures text][(g:300, tdim:3, tdel:2)]{
\includegraphics[width=0.34\textwidth]{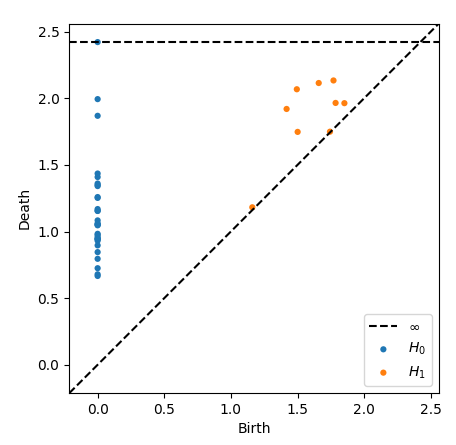}
\label{fig:subfig6r}}\\[-2ex]
\subfloat[Subfigure 4 list of figures text][(g:100, tdim:2, tdel:3)]{
\hspace*{-1.9em}
\includegraphics[width=0.34\textwidth]{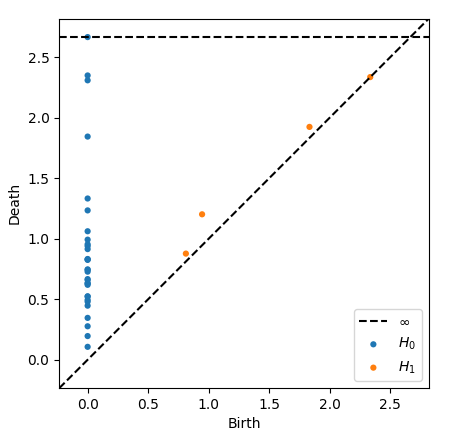}
\label{fig:subfig7r}} 
\subfloat[Subfigure 4 list of figures text][(g:200, tdim:2, tdel:3)]{
\includegraphics[width=0.34\textwidth]{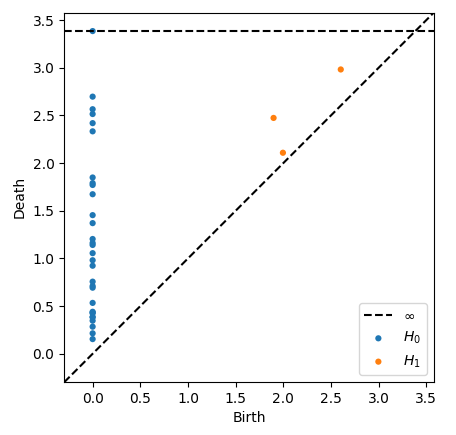}
\label{fig:subfig8r}} 
\subfloat[Subfigure 4 list of figures text][(g:300, tdim:2, tdel:3)]{
\includegraphics[width=0.34\textwidth]{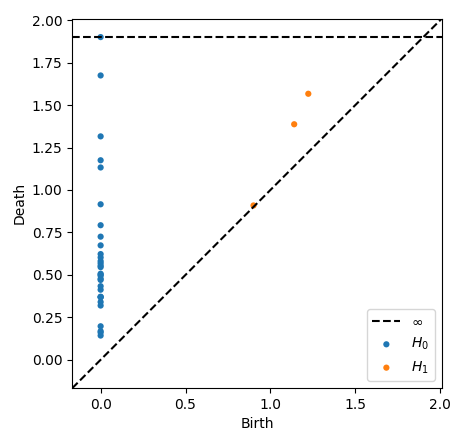}
\label{fig:subfig9r}}
\caption{\small{This figure represents persistence diagrams for the random text example from \cite{tymochko2020argumentative}: \textit{``The body may perhaps compensates for the loss of a true metaphysics. Yeah, I think it's a good environment for learning English. Wednesday is hump day, but has anyone asked the camel..."} \ The first parameter is the dimension of the (Glove) word embedding  vectors; the second parameter is the dimension of Takens embedding; and the third parameter is the delay of Takens embedding.  } }
\label{fig:randomText}
\end{figure}

\FloatBarrier

\subsection{Random text}\label{sec:randomText}
The cited work uses the following example to argue that random text produces chaotic display of persistence:
\textit{The body may perhaps compensates for the loss of a true metaphysics. Yeah, I think it's a good environment for learning English. Wednesday is hump day, but has anyone asked the camel...}

As shown in Figure~\ref{fig:randomText}, there seem to be no discernible pattern in persistence diagrams for this text. Depending on the choice of parameters we see that the same random text can produce virtually no signal, as in panel (b); produce `random' signal, as in panel (d), produce relatively strong signal in panels (c) and (g); and some patterns in between in the remaining panels.

\section{Discussion/Conclusion}\label{sec:discussion}

The first point we want to make is that the examples shown in the previous section show that there \textit{seem} to be no \textit{clear} relationship between persistence and circularity of a textual argument, contrary to the hypothesis proposed in \cite{tymochko2020argumentative}. Neither, there \textit{seem} to be such a relationship with respect to random text. 

The second point is that our counter-examples do not \textit{ prove} that there are no significant relations between topological features and circularity. An even stronger point we want to make is that our intuitions agree with \cite{tymochko2020argumentative}, and that such relationships are plausible. We hope and expect them to be discovered eventually, but we feel that the methods will likely to be more subtle. Our mathematical intuition points to the fact that topology and logic are connected through Heyting/Brouwer algebras (see e.g. \cite{rasiowa1963mathematics,sikorski1962applications}). On the TDA side, our earlier and current research \cite{gholizadeh2018topological,savle2019topological,
gholizadeh2020novel,gholizadeh2020topological} suggests that the contribution of topology to classification and/or inference may lie in augmenting other methods (and not necessarily being the center of the show as in  \cite{doshi2018movie}). In particular, it is possible we could see some statistical dependencies between persistence and circularity, if we ever run proper testing on a dataset of circular and non-circular arguments. Alas, we do not think such a dataset exists. Thus, we end this short note by pointing to these two avenues of exploration, one mathematical and one experimental.\\

\textbf{Note:} The references below are very incomplete, and we suggest the readers consult the list of references in \cite{tymochko2020argumentative} for a more appropriate introduction to topological data analysis. 

\bibliographystyle{plain}
\bibliography{ArgumentativeTopology}

\newpage
\section*{Code Appendix}

This code runs on Google Colab under the assumption that Glove vectors are installed in a particular directory. Therefore, to run it, the paths to word embeddings should be appropriately changed. 

The next four figures have all the commands required to replicate or extend our results. However, the code for entities with a $200$ in their names should be replicated (and adjusted) for different size embedding vectors. For dimension 200 the code runs in exactly in its current form.\\

\begin{figure}[h]
\centering

\hspace*{-2.9em}
\includegraphics[width=1.14\textwidth]{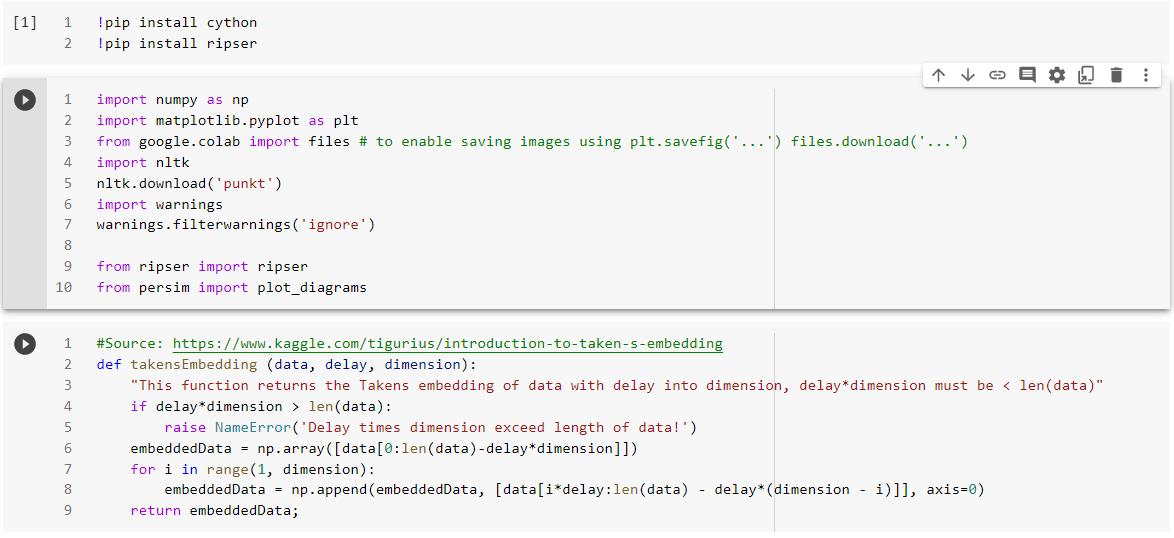}
\caption{\small{\textbf{Step 1.} First, we install the preliminaries, and then the code which will be used to produce Takens embeddings. The source of this function is given in the comment.}}
\label{fig:code0}
\end{figure}

\begin{figure}[t]
\centering

\hspace*{-2.9em}
\includegraphics[width=1.14\textwidth]{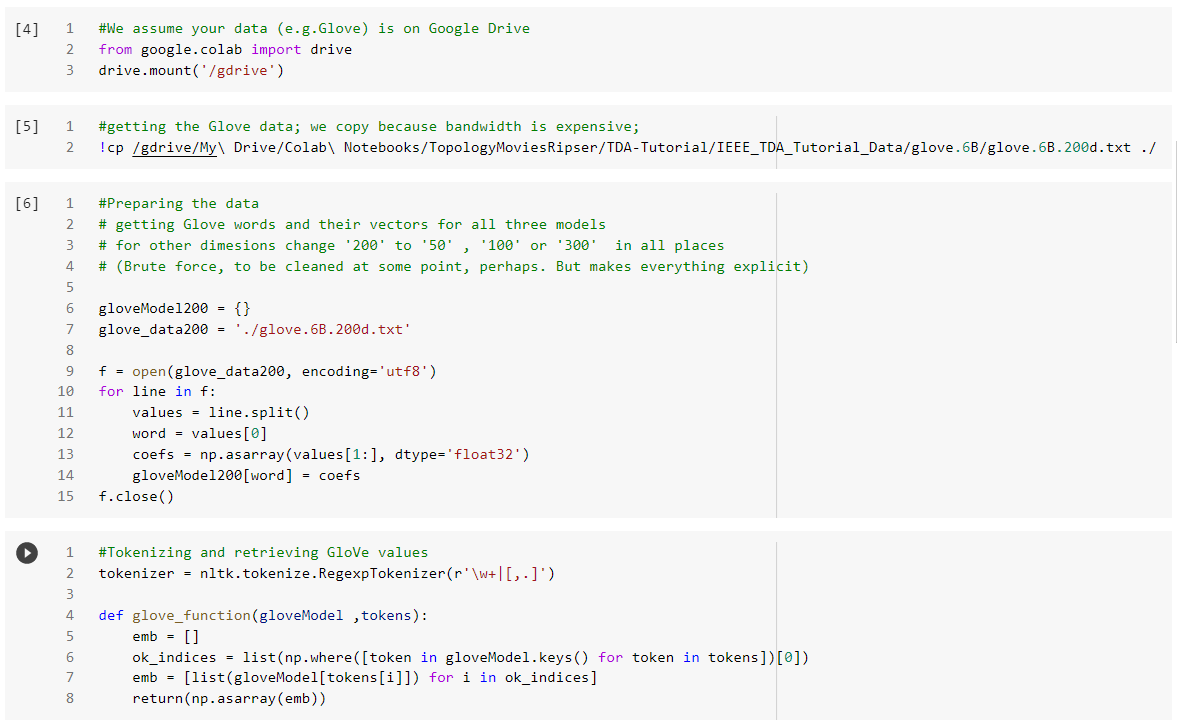}
\caption{ \small{\textbf{Step 2.} Data preparation. Getting the embedding vectors, creating a map between words and vectors. (Models for other dimensions are created in exactly the same way). The method \texttt{glove\_function} maps tokens into embedding vectors.}}
\label{fig:code1}
\end{figure}

\begin{figure}[t]
\centering

\hspace*{-2.9em}
\includegraphics[width=1.14\textwidth]{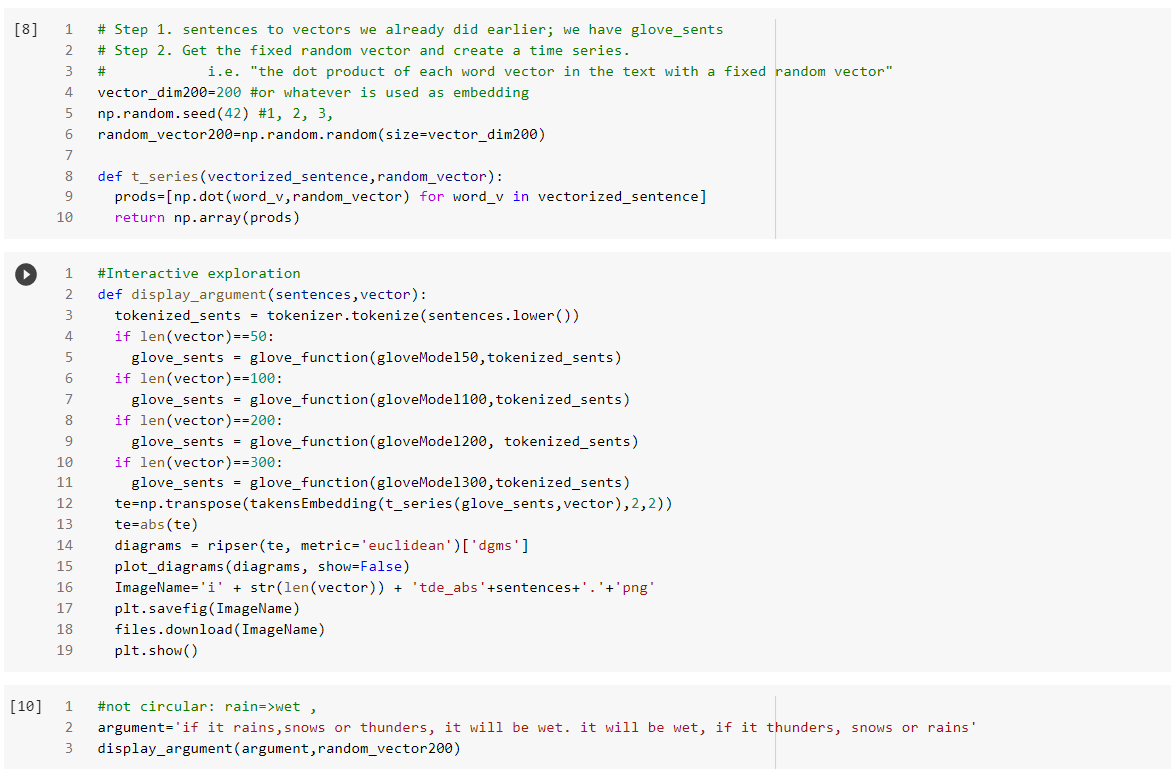}
\caption{\small{\textbf{Step 3.} The times series are obtained by projections on a random vector (as in \cite{tymochko2020argumentative}). The next panel allows an interactive exploration of persistence diagrams for different arguments, and the saving of the images. An example of using it is shown (without the output). }}
\label{fig:code2}
\end{figure}

\begin{figure}[t]
\centering

\hspace*{-2.9em}
\includegraphics[width=1.14\textwidth]{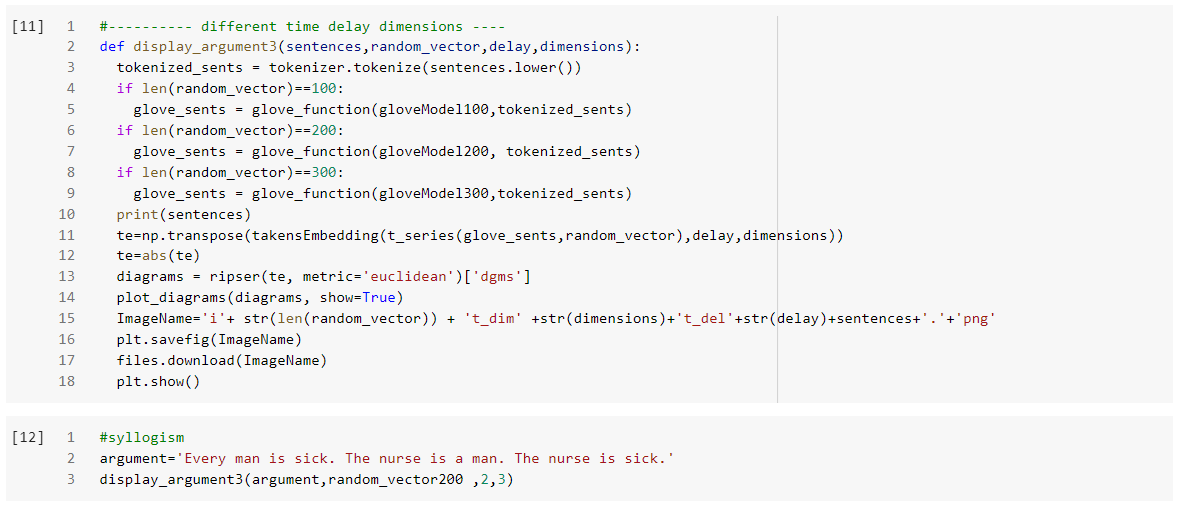}
\caption{\small{\textbf{Optional:} The method \texttt{display\_argument3} enables the user to interactively explore persistence diagrams with arbitrary parameters of Takens embeddings.}}
\label{fig:code3}
\end{figure} 
\end{document}